\PassOptionsToPackage{pdftex,dvipsnames,table}{xcolor}
\PassOptionsToPackage{export}{adjustbox}
\documentclass[lettersize,journal]{IEEEtran}

\usepackage{orcidlink}
\usepackage{amsmath,amsfonts}
\usepackage{algorithmic}
\usepackage{algorithm}
\usepackage{hyperref}
\usepackage{array}
\usepackage[caption=false,font=normalsize,labelfont=sf,textfont=sf]{subfig}
\usepackage{textcomp}
\usepackage{stfloats}
\usepackage[export]{adjustbox}
\usepackage{url}
\usepackage{booktabs}
\usepackage{verbatim}
\usepackage{graphicx}
\usepackage{cite}
\usepackage{comment}
\usepackage{bbm}
\usepackage{multirow} 

\usepackage{xargs}                      

\usepackage[export]{adjustbox}
\usepackage{mathtools}
\DeclarePairedDelimiter{\ceil}{\lceil}{\rceil}

\definecolor{RoyalBlue}{HTML}{4169e1}
\definecolor{ForestGreen}{HTML}{228b22}

\graphicspath{{./}{figures/}}

\def\BibTeX{{\rm B\kern-.05em{\sc i\kern-.025em b}\kern-.08em
    T\kern-.1667em\lower.7ex\hbox{E}\kern-.125emX}}

\usepackage{balance}

\begin{document}


\title{Beyond MMD: Evaluating Graph Generative Models with Geometric Deep Learning\\
 }
 \author{%
     \IEEEauthorblockN{1\textsuperscript{st} Salvatore Romano}
     \IEEEauthorblockA{\textit{Dipartimento di Ingegneria Elettrica, Elettronica e Informatica}\\
     \textit{University of Catania, Italy} \\
     salvatore.romano@phd.unict
      \orcidicon{}
     }
     \and
     \IEEEauthorblockN{1\textsuperscript{st} Marco Grassia}
    \IEEEauthorblockA{\textit{Dipartimento di Ingegneria Elettrica, Elettronica e Informatica}\\
    \textit{University of Catania, Italy} \\
     marco.grassia@unict.it
      \orcidicon{0000-0001-5841-6058}
     }
    \and
     \IEEEauthorblockN{4\textsuperscript{th} Giuseppe Mangioni}
     \IEEEauthorblockA{\textit{Dipartimento di Ingegneria Elettrica, Elettronica e Informatica}\\
     \textit{University of Catania, Italy} \\
     giuseppe.mangioni@unict.it
      \orcidicon{0000-0001-6910-0112}
     }
 }
\author{%
Salvatore Romano~\orcidlink{0009-0002-2550-4435},
Marco Grassia~\orcidlink{0000-0001-5841-6058},
Giuseppe Mangioni~\orcidlink{0000-0001-6910-0112} \\
\thanks{%
S. Romano, M. Grassia, and G. Mangioni are with the Department of Electrical, Electronic and Computer Engineering (DIEEI), University of Catania, Catania, Italy.
S. Romano is also with the University Campus Bio-Medico di Roma, Rome, Italy.
e-mail: salvatore.romano@phd.unict.it; marco.grassia@unict.it; giuseppe.mangioni@unict.it
}
}

\markboth{Journal of \LaTeX\ Class Files,~Vol.~18, No.~9, September~2020}%
{Romano \MakeLowercase{\textit{et al.}}: Beyond MMD: Evaluating Graph Generative Models with Geometric Deep Learning}


\maketitle

\begin{abstract}
Graph generation is a crucial task in many fields, including network science and bioinformatics, as it enables the creation of synthetic graphs that mimic the properties of real-world networks for various applications.
Graph Generative Models (GGMs) have emerged as a promising solution to this problem, leveraging deep learning techniques to learn the underlying distribution of real-world graphs and generate new samples that closely resemble them.
Examples include approaches based on Variational Auto-Encoders, Recurrent Neural Networks, and more recently, diffusion-based models. 
However, the main limitation often lies in the evaluation process, which typically relies on Maximum Mean Discrepancy (MMD) as a metric to assess the distribution of graph properties in the generated ensemble.
This paper introduces a novel methodology for evaluating GGMs that overcomes the limitations of MMD, which we call RGM (Representation-aware Graph-generation Model evaluation).
As a practical demonstration of our methodology, we present a comprehensive evaluation of two state-of-the-art Graph Generative Models: Graph Recurrent Attention Networks (GRAN) and Efficient and Degree-guided graph GEnerative model (EDGE).
We investigate their performance in generating realistic graphs and compare them using a Geometric Deep Learning model trained on a custom dataset of synthetic and real-world graphs, specifically designed for graph classification tasks.
Our findings reveal that while both models can generate graphs with certain topological properties, they exhibit significant limitations in preserving the structural characteristics that distinguish different graph domains.
We also highlight the inadequacy of Maximum Mean Discrepancy as an evaluation metric for GGMs and suggest alternative approaches for future research.
\end{abstract}

\begin{IEEEkeywords}
Graph generation, graph neural networks, geometric deep learning, maximum mean discrepancy, evaluation metrics, siamese networks, graph classification.
\end{IEEEkeywords}

\section{Introduction}
\IEEEPARstart{C}{\lowercase{omplex}} systems, characterized by many non-trivially interacting components, are pervasive in nature, society, and technology and often exhibit emergent phenomena, self-organization, and complex dynamics.
Researchers often represent them as networks -- graphs in the mathematical sense -- where nodes (vertices) represent entities and links (edges) represent relationships between them. Due to their characteristics, these networks are often referred to as complex networks, and a new field of research, called Network Science, has emerged to study them.
To understand and analyze complex networks and the underlying systems, researchers rely on mathematical, physical and computational models, statistics, and machine learning techniques.
For instance, the degree distribution, clustering coefficient, and average path length are some of the most common topological properties used to characterize complex networks, while the presence of motifs, community structure, hierarchical organization, and assortativity (i.e., the tendency of nodes to connect to similar nodes, homophily in jargon) are also important features that can provide insights into the underlying processes and mechanisms that govern the network's behavior~\cite{newman2018networks}.
Moreover, robustness, resilience, and vulnerability are also important properties to consider when analyzing complex networks, as they can provide insights into the network's ability to withstand perturbations or attacks~\cite{grassia2024robustness,grassia2021machine}.
An open key question in Network Science is how to model real-world networks that capture all these properties, as they are often hard to quantify and reproduce.
In fact, the ability to generate realistic networks is crucial in many fields including bioinformatics, social sciences, (network) medicine, and others, since it enables the evaluation of graph algorithms, the study of network dynamics, and the generation of null models. 

Graph generation is the process of creating synthetic graphs that have desired properties, ultimately mimicking the characteristics of real-world networks.
While a plethora of non-trainable graph generation models have been proposed, such as the Erdős-Rényi model~\cite{erdos59a}, the Barabási-Albert model~\cite{Baraba_si_1999}, and the Stochastic Block Model~\cite{HOLLAND1983109}, they reflect only a limited set of graph properties --- such as degree distribution or clustering coefficient, or a trivial combination of them --- that are insufficient to capture the complexity of real-world graphs and their underlying systems. This often results in unrealistic structures with limited applicability.
For example, the Erdős-Rényi model generates graphs with a Poisson degree distribution, while the Barabási-Albert model produces scale-free networks with a power-law degree distribution thanks to its preferential attachment mechanism. This process favors the growth of hubs and leads to the emergence of a few highly connected nodes due to the ``rich get richer'' phenomenon.
However, these models fail to capture other important properties of real-world networks, such as community structure, clustering, and the presence of motifs.
The Stochastic Block Model can generate graphs with a predefined community structure, but it often relies on strong assumptions about the underlying graph structure and the number of communities, which may not be fixed in real-world networks.
This limitation is partially overcome, for instance, by the LFR (Lancichinetti-Fortunato-Radicchi) benchmark~\cite{Lancichinetti_2008}, a widely used synthetic graph generator that creates graphs with a power-law degree and community distributions, but it is still limited in its ability to capture other important properties, such as clustering and motifs.
Other generative models, such as the nPSO (non-uniform Popularity-Similarity Optimization) model~\cite{Muscoloni_2018}, sample from a non-uniform distribution of angular coordinates in the hyperbolic disk --- typically a Gaussian mixture distribution --- to create agglomerates of nodes that are more densely connected within their communities than with the rest of the network, and also allow controlling the clustering coefficient.
However, all these algorithms inherently encode human assumptions about the underlying graph structure, which might work well for specific applications but fail to generalize across different domains, making them inadequate for applications that demand a more realistic representation of complex systems.
On the other hand, in the last decade, the use of machine learning (and deep learning) has revolutionized many domains, including computer vision, natural language processing, and speech recognition, where deep learning techniques have proven effective in capturing the complex and diverse characteristics of real-world data.
For instance, deep learning has been successfully used in addressing open questions in complex systems like the prediction of protein structures~\cite{article}, the generation of realistic images~\cite{goodfellow2014generativeadversarialnetworks}, and approaching combinatorial and NP--hard problems~\cite{10.1145/3551642,10.1371/journal.pone.0296185}.
In the context of graph generation, Graph Generative Models (GGMs) have emerged as a promising data-driven solution to this problem, leveraging deep learning techniques to learn the underlying distribution of real-world graphs and generate new samples that closely resemble them.
Examples of GGMs include Variational Graph Autoencoders (VGAE)~\cite{kipf2016variationalgraphautoencoders}, Graph Recurrent Attention Networks (GRAN), and more recently diffusion-based models like ``Efficient and Degree-guided graph Generative model'' (EDGE).
However, the main limitation of these models often lies in the evaluation process, which typically relies on Maximum Mean Discrepancy (MMD) as a metric, which is not suitable for properly assessing the quality of generated graphs.

Our work introduces a novel methodology for the evaluation of GGMs that overcomes the limitations of MMD, which we call RGM (Representation-aware Graph-generation Model evaluation).
As a practical demonstration of our methodology, we present a comprehensive evaluation of two state-of-the-art Graph Generative Models: GRAN and EDGE.
We investigate their performance in generating realistic graphs and compare them using a Geometric Deep Learning model trained on a custom dataset of synthetic and real-world graphs, specifically designed for graph classification tasks where the goal is to distinguish between different graph domains.
In fact, when evaluating the performance of generative models, it is crucial to capture the underlying patterns and relationships present in the original dataset. 
Our findings reveal that while both models can generate graphs with specific topological properties, they exhibit significant limitations in preserving structural properties that distinguish different graph domains.
We also highlight the inadequacy of Maximum Mean Discrepancy (MMD) as an evaluation metric for GGMs, suggesting alternative approaches for future research.

\section{Related Works}
\subsection{Graph Generation}
A variety of heuristic graph generation models have been proposed to create synthetic graphs that exhibit specific properties or structures.
In fact, while generating a synthetic graph is relatively straightforward, ensuring that the generated graph accurately reflects specific properties or resembles the characteristics of real-world networks is arguably a challenging task.
The proposed heuristic models often rely on human assumptions about the underlying graph structure, which may work well for specific applications but fail to generalize across different domains.
Moreover, while most of these models can generate the topological properties of the graph, they often fail to capture other important properties that are crucial for a realistic representation of a complex system.
For instance, domain-specific topological properties, such as hierarchical organization, robustness, and motifs, or even the feature distribution of the nodes or edges, are often overlooked.

Here we present a brief overview of some heuristic graph generation models that have been proposed in the graph theory and network science literature for a wide variety of applications, including but not limited to, analytical graph characterization, community detection and link prediction.
\paragraph{Erdős-Rényi (ER) model}
The ER model~\cite{erdos59a} is the first ever proposed algorithm for the formation of random graphs.
In its standard form, denoted as $G(n,p)$, it generates a graph with $n$ nodes where each possible edge between a pair of nodes is independently included with probability $p$.
This model leads to graphs that are typically homogeneous, with node degrees concentrated around their mean, and no inherent geometric or community structure.
ER graphs do not exhibit community structure and have a homogeneous degree distribution (meaning most nodes have similar degrees), which results in robustness to random failures and to targeted attacks.
Yet, the Erdős-Rényi model is well known for its simplicity and efficiency in generating random graphs, has been widely studied, and serves as a foundational model in graph theory.
For such reasons, it is often used, for instance as a null model for comparison with real-world networks, as it provides a baseline for understanding the significance of observed network properties.

\paragraph{Stochastic Block Model (SBM)}
The SBM~\cite{HOLLAND1983109} is a generative model for random graphs that captures community structure by dividing nodes into blocks/groups and assigning edge probabilities based on block memberships.
That is, the probability of an edge between two nodes depends on the blocks to which they belong.
Formally, given a set of nodes partitioned into $K$ blocks, edges between node pairs are drawn independently according to a block-dependent probability matrix $P \in [0,1]^{K \times K}$, where $P_{ij}$ defines the probability of an edge between a node in block $i$ and a node in block $j$.
It can be viewed as a generalization of the Erdős-Rényi model, where the probability of edge formation is not uniform across all nodes.
The SBM model is particularly useful for modeling networks with community structure, where nodes within the same block are more likely to be connected than nodes in different blocks, but it suffers from limitations such as the need to specify the number of blocks and their sizes in advance, which may not be known in practice.

\paragraph{Barabási-Albert (BA) model}
The Barabási–Albert model~\cite{Baraba_si_1999} constructs a growing graph by simulating the process of preferential attachment, where new nodes are added to the graph and preferentially attach to existing nodes with a probability proportional to their degree, giving more connected nodes a higher chance of receiving new links in a ``rich-get-richer'' fashion.
In particular, the model starts with a small number of nodes connected with arbitrary topology (often a fully connected graph), and iteratively adds new nodes, where each new node connects to a fixed number $m$ of existing nodes with probability proportional to their degrees.
Formally, the probability $P_{i}$ that the new node is connected to node $i$ is:
\begin{equation}
    P_i = \frac{k_i}{\sum_j k_j}
\end{equation}
where $k_i$ is the degree of node $i$ and the sum is over all pre-existing nodes $j$.
While the BA model generates a power-law degree distribution often exhibited by real-world networks, it fails to capture other important properties such as community structure, target clustering coefficients, or specific motifs.
\paragraph{Lancichinetti-Fortunato-Radicchi (LFR) model}
The LFR benchmark~\cite{Lancichinetti_2008} is a widely used synthetic graph generator that creates graphs with a power-law degree and community distributions.
It often serves as a benchmark for testing and comparing community detection algorithms, as it allows researchers to generate networks with known community structures and evaluate the performance of their algorithms in detecting these communities.
The LFR model includes a mixing parameter $\mu$ that controls the fraction of edges that are between communities, reflecting the amount of ``noise'' in the network: $\mu=0$ indicates no inter-community edges and $\mu=1$ indicates all edges are inter-community.
The graph is generated by sampling a degree sequence from a power-law distribution with a specified exponent, assigning nodes to communities such that each community size follows another power-law distribution, while also considering the limitations imposed by the mixing parameter $\mu$, and finally adding links between nodes based on their community memberships and the mixing parameter.

\paragraph{non-uniform Popularity-Similarity-Optimization (nPSO) model} 

The nPSO~\cite{Muscoloni_2018} model is designed to generate complex networks with realistic community structures by addressing the limitations of previous models like PSO~\cite{Papadopoulos_2012} and GPA~\cite{Zuev_2015}.
In the original Popularity-Similarity-Optimization (PSO) model, a graph is generated by simulating its growth process in a hyperbolic space, where nodes are placed in a polar coordinate system and connected based on their distance from each other.
Specifically, two factors are considered: \textit{Popularity}, which is represented by the radial coordinate of the nodes, and \textit{Similarity}, represented by the angular coordinate.
During growth, new nodes are added to the graph and connected to a fixed number of existing nodes based on their distance in the hyperbolic space, which is iteratively increased to simulate the fading of popularity.
The Geometrical-Preferential-Attachment (GPA) model builds on the PSO model by introducing a third factor, called \textit{Attractiveness}, a form of preferential attachment based on hyperbolic distance between nodes, that determines the probability of connecting to existing nodes and influences the growth process and the resulting graph structure, which exhibits scale-free degree distribution and a non-trivial community structure.
nPSO further refines the model by sampling from a non-uniform distribution of nodes over the hyperbolic disk, which allows for the creation of agglomerates of nodes that are more densely connected within their communities than with the rest of the network.
For instance, if a Gaussian mixture distribution is used, the resulting graph will exhibit communities that emerge in correspondence with the different Gaussians,
allowing the control of the number of communities and their sizes.
\subsection{Graph Classification and Graph Similarity}
Graph classification is the task of assigning a label to a graph based on its structure and properties.
Formally, given a set of graphs $G=\{G_1, \ldots, G_N\}$ and their labels $y = \{y_1, \ldots, y_N\}$, the goal is to predict the label $y_i$ of each graph $G_i$.
Many approaches have been proposed, including traditional methods that are based on hand-crafted feature extraction and more recent deep learning-based ones.
For instance, traditional methods often rely on graph kernels, which first extract a set of hand-crafted features (e.g., node degree, clustering coefficient, node coloring, etc.) and then use a kernel function to measure the similarity between graphs based on these features.
On the other hand, deep learning-based methods leverage neural networks to learn representations of graphs, allowing for more flexible and powerful classification models, which may not involve hand-crafted features or some possibly biased assumptions about the graph structure, as it learns the features directly from the data.
In this context, Graph Representation Learning (GRL), also known as Geometric Deep Learning~\cite{DBLP:journals/corr/BronsteinBLSV16}, has emerged as a powerful framework for learning representations of graph-structured data.
The basic idea is that --- as common in most deep learning approaches --- the model learns a mapping function from the input space (i.e., the graph) to a high-dimensional embedding space (i.e., a latent space where a graph $G_i$ is represented as a vector $h_{G_i}$) using a neural network.
Such representation should preserve some useful properties of the original input data, and should be useful for the downstream task like, but not limited to, graph classification.


The most popular approach to learn representations of graphs are Graph Neural Networks (GNNs)~\cite{hamilton2020graph,bronstein2021geometric,4700287}, based on a message-passing mechanism that allows each node to exchange information with its neighbors and to build its own representation based on a trainable function of the information received from its neighbors.
That is, GNNs iteratively update the representation of each node by aggregating incoming messages from its neighbors, which is, in turn, sent to its own neighbors in the next iteration to update their representations.
A plethora of GNNs have been proposed in the literature, each with its own architecture and design choices.
For instance, Graph Convolutional Networks (GCNs)~\cite{kipf2017semisupervisedclassificationgraphconvolutional} use a convolutional operator (i.e., a shared filter) to aggregate information from neighboring nodes, GraphSAGE~\cite{hamilton2018inductiverepresentationlearninglarge} extends this idea by using a sampling strategy and sampling a fixed-size neighborhood to allow the model to scale to large graphs, while Graph Attention Networks (GATs)~\cite{veličković2018graphattentionnetworks,grassia2021wsgat} use an attention mechanism to weigh the importance of neighboring nodes when aggregating their features, allowing the model to focus on the most relevant ones for the task, and Graph Isomorphism Networks (GIN)~\cite{DBLP:journals/corr/abs-1810-00826} are inspired by the Weisfeiler-Lehman graph isomorphism test and use a neural network (e.g., a multi-layer perceptron) to learn a representation of the graph based on its structure.

From node representations, graph-level representations (i.e., the graph embedding) can be obtained by pooling the representation of its most elementary components (i.e., nodes) using a permutation-invariant pooling function like global pooling functions (e.g., mean, max, sum) or by leveraging more sophisticated techniques such as attention mechanisms or hierarchical aggregation~\cite{ying2019hierarchicalgraphrepresentationlearning,ranjan2020asapadaptivestructureaware}.
Specifically, a Geometric Deep Learning (GDL) model is usually composed of a series of GNN layers, interspersed with non-linear activation functions, followed by the pooling layer.
That is, $h_G = \operatorname{pool}(h_{v_1}, h_{v_2}, \ldots, h_{v_n})$, where $h_G$ is the graph-level representation, $h_{v_i}$ is the representation of node $v_i$, and $\operatorname{pool}$ is a permutation-invariant pooling function.

Once the graph-level representation is obtained, it can be used for various tasks (including, of course, graph classification) using classic ML techniques, such as fully connected layers, support vector machines, or other classifiers.
In this sense, similarity between two graphs can be defined as the distance between their respective representations in the embedding space, which is often learned by the GDL model for a specific task, such as graph classification.
More formally, given two graphs $G_i$ and $G_j$, they are similar if their representations $h_{G_i}$ and $h_{G_j}$ are close in the embedding space.
In this context, graph similarity computation~\cite{jin2022cgmncontrastivegraphmatching,DBLP:journals/corr/abs-1912-11615} has seen significant advancements in recent years and has proven to be highly beneficial across various domains, but has not yet been tested in the field of graph generation extensively.
A method for improving graph generation evaluation using a classifier was proposed in the literature by Liu et al.~\cite{Liu2019AutoregressiveGG}, although it was demonstrated on a limited dataset and with restricted use of Graph Generative Models.
Specifically, they trained both the Graph Generative Models and a Graph Isomorphism Network (GIN) classifier on small graph datasets such as \textit{Community-small}, \textit{Grid-small}, and \textit{Ego-small}.
While this work represents a step forward, it has several limitations, including the use of small datasets and the limited evaluation of the generated graphs.
Here, we address these limitations by proposing a more comprehensive evaluation methodology that leverages larger datasets that include both real-world and synthetic graphs, by employing a more sophisticated model architecture and training methodology, and by providing a more thorough evaluation of the generated graphs.

\subsection{Graph Generative Models (GGMs)}

Graph generation using Graph Generative Models (GGMs) is a machine learning task focused on capturing the underlying patterns, structures, and properties of a given set of graph data to generate new graphs that closely resemble the originals.
Unlike traditional approaches, their primary goal is to use the learned latent representation of the input graphs to generate new ones that reflect the characteristics of the original dataset.

The early development of GGMs leveraging deep learning techniques was spearheaded by GraphVAE~\cite{simonovsky2018graphvaegenerationsmallgraphs}, which laid the groundwork for subsequent advancements in the field.
GGMs leverage deep learning to capture complex relationships and dependencies between nodes and edges. This allows them to model intricate structural properties and, ideally, generate graphs that closely reflect the topology and statistical characteristics of the original dataset.
As a result, GGMs are a powerful tool for graph generation tasks.

More formally, the objective in graph generation~\cite{guo2022systematicsurveydeepgenerative} is to learn a generative model capable of producing graphs that closely resemble a given set of input graphs $\{G^{(i)}\}$.
This involves modeling the underlying data distribution and ensuring the ability to sample realistic graphs. Specifically:
\begin{itemize}
    \item $p_{\text{data}}(G)$: represents the true data distribution from which the graphs are sampled, i.e., $G^{(i)} \sim p_{\text{data}}(G)$, where $\sim$ means sampled or drawn.
    \item $p_\theta(G)$: represents the generative model, parameterized by $\theta$, which is designed to approximate $p_{\text{data}}(G)$.
\end{itemize}
The goal is twofold:
\begin{enumerate}
    \item \label{firstgoal} Ensure that $p_\theta(G)$ closely approximates $p_{\text{data}}(G)$.
    \item \label{secondgoal} Enable efficient sampling from $p_\theta(G)$ to generate realistic graphs.
\end{enumerate}

We can employ an auto-regressive model that satisfies these two goals:
\begin{equation}
    p_\theta(G^{(i)}) = \prod_{t=1}^{T} p_\theta(A^{(i)}_t \mid A^{(i)}_1, \ldots, A^{(i)}_{t-1})
\end{equation}
where $T$ is the total number of generation steps, and $A^{(i)}_t$ represents the adjacency matrix of $G^{(i)}$ after step $t$ of the generation process, which may involve adding a node, adding an edge, or more generally modifying existing connections.
However, as the complexity of the graph increases, the evaluation of the generative process becomes more challenging and must consider various structural aspects and properties, such as degree distribution, clustering coefficient, community structure, motif prevalence, and connectivity patterns, which are essential for accurately reflecting the characteristics of real-world graphs.
While some approaches rely on visual inspection of the generated graphs, we argue that this is not a reliable method for evaluating the performance of GGMs.
Most recent approaches rely on Maximum Mean Discrepancy (MMD) as a metric to assess the distribution of graph properties in the generated ensemble.

In the following, two state-of-the-art GGM approaches are introduced.
\paragraph{GRAN}
\textit{Graph Recurrent Attention Network} (GRAN)~\cite{liao2020efficientgraphgenerationgraph} is an auto-regressive graph generative model that leverages the attention mechanism and Graph Neural Networks (GNNs) to generate graphs sequentially, simulating a growth process.
The model generates blocks of nodes and their connections to the existing graph at each step. 

Let $G=(V,E)$ be an undirected, unweighted graph with node set $V$ and edge set $E$.
GRAN models the distribution over graphs via their adjacency matrices $A^\pi$, where $\pi$ denotes a particular node ordering.
Unlike traditional autoregressive approaches that sequentially generate each edge or node, GRAN adopts a block-wise decoding strategy.
The adjacency matrix $A^\pi$ is generated row-wise in blocks of size $B$.
Without loss of generality, we can assume that the graph is undirected and that the lower triangle $L^\pi$ is generated and $A$ is symmetric.
If that's not the case, it can be easily achieved by generating the upper triangle in the same way.
The model generates one block of $B$ rows of $L^\pi$ at a time, thus the number of steps to generate a graph is given by $T = \ceil*{N/B}$, where $N = |V|$.
The joint probability of connections in the graph $G$ is factorized as:
\begin{equation}
    p_\theta(L^\pi) = \prod_{t=1}^T p_\theta(L_{b_t}^\pi \mid L_{b_1}^\pi, \ldots, L_{b_{t-1}}^\pi)
\end{equation}
where $L_{b_t}^\pi$ denotes the $t$-th block of $B$ rows in the lower triangular adjacency matrix $L^\pi$.
At each generation step $t$, GRAN constructs an augmented graph called $G_t$, which includes the already generated nodes $\{b_1,\ldots,b_{t-1}\}$ and the current target block $b_t$, aiming at predicting the connectivity of $b_t$.
Each existing node $i$ is initialized with a learned embedding based on its corresponding row in the partial adjacency matrix:
\begin{equation}
    h^0_{i} = W L_{i}^\pi + b,
\end{equation}
where $L_{i}^\pi$ is the $i$-th row of the lower triangular matrix generated so far, while embeddings for newly added nodes in block $b_t$ are initialized to zero.
In the architecture, a fixed number $R$ (rounds) of GNN message-passing layers are then applied.
In each layer, messages are computed as: 
\begin{equation}
    m^r_{ij} = f(h_i^r - h_j^r),
\end{equation}
with attention coefficient determined by: 
\begin{equation}
    a_{ij}^r = \sigma(g(\Tilde{h}^r_i-\Tilde{h}_j^r)), 
    \Tilde{h}_i^r = [h_i^r, c_i]
\end{equation}
where $c_i$ is an encoding of node types (existing vs newly added), $f$ and $g$ are learnable neural networks, and $\sigma$ is the sigmoid activation.
Updated node representations are computed using a Gated Recurrent Unit (GRU) as in~\cite{li2017gatedgraphsequenceneural}.
GRAN utilizes a mixture of Bernoulli distributions to model the probability of generating edges in the block $L_{b_t}^\pi$ by using the final node embeddings $h_i^R$ for each node $i$.
Formally we have:
\begin{equation}
    p_\theta(L_{b_t}^\pi \mid L_{b_1}^\pi, \ldots, L_{b_{t-1}}^\pi) = \sum_{\ell=1}^K \alpha_\ell \prod_{i \in b_t} \prod_{1 \leq j \leq i} \rho_{\ell,i,j}
\end{equation}
where $K$ is the number of mixture components; $\alpha_\ell$ are the mixture weights that influence the edge generation and are computed using an MLP applied to node embedding differences.
The term $\rho_{\ell,i,j} \in [0,1]$ denotes the probability that an edge exists between nodes $i$ and $j$ under mixture component $\ell$, and is obtained via an MLP followed by a sigmoid activation applied to the difference between node embeddings $h_i^R$ and $h_j^R$.

As an autoregressive model, GRAN has certain limitations, such as sequential generation and node ordering, which can restrict its ability to capture the global structure of the graph.
Unlike GraphRNN, which relies on RNNs and carries hidden states during generation, GRAN uses GNNs to ensure that the generation decisions for a particular block $b_t$ are influenced by the graph's structure rather than relying on sequential hidden states. 

\paragraph{EDGE}
\textit{Efficient and Degree-Guided Graph Generative model} (EDGE)~\cite{chen2023efficientdegreeguidedgraphgeneration} is a diffusion-based graph generative model.
Like other diffusion models, EDGE is trained by learning to reverse a forward noising process. During training, a forward Markov chain progressively corrupts graphs by removing edges; during generation, a learned denoising model inverts this process, starting from an empty graph and adding edges progressively until a realistic graph is produced.

Following the standard diffusion model convention, let $\mathbf{A}_0$ denote the adjacency matrix of the target (clean) graph, and $\mathbf{A}_t$ the adjacency matrix at diffusion step $t$.
EDGE defines a forward Markov chain that progressively corrupts the graph:
\begin{equation}
    q(\mathbf{A}_{1:T}\mid \mathbf{A}_0) = \prod_{t=1}^{T} q(\mathbf{A}_t\mid \mathbf{A}_{t-1}),
\end{equation}
where each transition independently removes edges with probability $\beta_t$:
\begin{equation}
    (\mathbf{A}_t)_{ij} \sim \operatorname{Bern}\bigl((1-\beta_t)(\mathbf{A}_{t-1})_{ij}\bigr)
\end{equation}
At each step $t$, existing edges are stochastically deleted, ensuring that the forward process converges to the all-zero adjacency matrix (empty graph) as $t \to T$. The posterior of the forward transition can be computed as:
\begin{equation}
    q(\mathbf{A}_{t-1} \mid \mathbf{A}_t, \mathbf{A}_0) = \frac{q(\mathbf{A}_t \mid \mathbf{A}_{t-1})\, q(\mathbf{A}_{t-1} \mid \mathbf{A}_0)}{q(\mathbf{A}_t \mid \mathbf{A}_0)}
\end{equation}

Rather than modeling all potential edges, EDGE selects a small subset of \textit{active} nodes and only predicts edges among them. The active node indicator is defined as $s_i^t := \mathbbm{1}[(\mathbf{d}_{t-1})_i \neq (\mathbf{d}_t)_i]$, where $\mathbf{d}_t = \operatorname{deg}(\mathbf{A}_t)$ is the degree sequence at step $t$. Thus, $s_i^t$ flags nodes whose degree changed due to edge removal. At each step, the model first predicts which nodes change degree, then adds edges only between the active nodes.

To enable degree-guided generation, EDGE treats the target degree sequence $\mathbf{d}_0$ as a random variable: it first samples $\mathbf{d}_0 \sim p_{\theta}(\mathbf{d}_0)$, then generates $\mathbf{A}_0$ conditioned on $\mathbf{d}_0$. Formally:
\begin{equation}
    p_{\theta}(\mathbf{A}_0, \mathbf{d}_0) = p_{\theta}(\mathbf{d}_0)\, p_{\theta}(\mathbf{A}_0 \mid \mathbf{d}_0)
\end{equation}
The full denoising (reverse) model is:
\begin{equation}
    p_{\theta}(\mathbf{A}_{0:T}, s^{1:T}, \mathbf{d}_0) = p_{\theta}(\mathbf{d}_0)\, p_{\theta}(\mathbf{A}_{0:T}, s^{1:T} \mid \mathbf{d}_0)
\end{equation}
Degree guidance is enforced by masking node selection: nodes whose current degree $(\mathbf{d}_t)_i$ has already reached the target $(\mathbf{d}_0)_i$ cannot become active.

By combining edge-removal diffusion, active-node selection, and degree conditioning, EDGE can scale discrete diffusion graph generation to thousands of nodes while matching key global graph statistics.

\section{Problem Formulation}
The problem addressed in this paper concerns the evaluation of graph generation algorithms, with a focus on the generative process of Graph Generative Models (GGMs).
In fact, current techniques for evaluating GGMs often rely on Maximum Mean Discrepancy (MMD) as a metric to assess the distribution of some graph properties in the generated ensemble, or even on the visual inspection of the generated graphs.

\subsection{Evaluation Metric: Maximum Mean Discrepancy}
Maximum Mean Discrepancy (MMD)~\cite{gretton2008kernelmethodtwosampleproblem}~\cite{JMLR:v13:gretton12a} is the most commonly used metric for evaluating the generative process of Graph Generative Models (GGMs) in the literature.
It is a distance-measure between distributions $P(X)$ and $Q(Y)$ that is meant to capture the similarity between a specific property of the original and the generated graphs, $X$ and $Y$ respectively.
Specifically, once a feature-extractor function $\phi$ is defined --- mapping the graphs into a feature space ---  MMD computes the distance between the mean embeddings of the two distributions in a reproducing kernel Hilbert space (RKHS).
The kernel function $k$ is mathematically defined as $k(x, x') = \langle \phi(x), \phi(x') \rangle$, where $\phi(x) \in \mathcal{H}$.
For instance, $\phi$ may extract the degree distribution of the graphs, the clustering coefficient, or other graph-level features, which are then compared using a kernel function $k$ to measure the similarity between the feature representations of graphs in the feature space. 
Formally, given $n$ samples $X = \{x_1,\ldots,x_n\} \subseteq \mathcal{X} $ and $m$ samples
$Y = \{y_1, \ldots, y_m\} \subseteq \mathcal{X}$, the biased empirical estimate of the MMD between $X$ and $Y$ can be defined as~\cite{DBLP:journals/corr/abs-2106-01098}:

\begin{equation}
\begin{split}
    \operatorname{MMD}^2 (X, Y) := & \frac{1}{n^2} \sum_{i, j = 1}^n k(x_i, x_j) + \\
    & + \frac{1}{m^2} \sum_{i, j = 1}^m
k(y_i, y_j) - \frac{2}{n m} \sum_{i = 1}^n \sum_{j = 1}^m k(x_i, y_j)
\end{split}
\label{e:mmd}
\end{equation}
where $k$ is a positive definite kernel function, which is a measure of similarity between two samples.
That is, MMD accounts for the inner similarity between all the graph pairs in both the dataset and the generated graphs, and also for the cross-similarity between the two sets of graphs.
The lower the MMD, the better.

This definition can quantify how closely the generated graphs resemble the original ones as a whole, according to some descriptor $\phi$, but has significant limitations that make it unsuitable for evaluating GGMs.
First and foremost, the effectiveness of MMD heavily relies on the choice of the descriptor $\phi$ and kernel $k$ functions.
In addition, MMD primarily compares the average value of the extracted property across the two sets, and may fail to capture differences in the internal distribution of that property within individual graphs.
Secondly, as graphs are inherently complex structures characterized by intricate topological properties, optimizing one specific property may lead to the degradation of others.

This leads to a combination of MMD values that are hardly comparable and may not reflect the overall quality of the generated graphs.
Additionally, MMD is unbounded and sensitive to outliers, which can significantly affect the results. Moreover, evaluating a single graph poses an additional challenge, as MMD is fundamentally designed to compare distributions rather than individual instances.
We refer the reader to~\cite{DBLP:journals/corr/abs-2106-01098,shirzad2022evaluating} for a more detailed discussion on the limitations of MMD and its application to graph generative models.

\subsection{Neural Graph Similarity}
Here, we propose RGM (Representation-aware Graph-generation Model evaluation), a novel approach for evaluating generated graphs by leveraging learned graph similarity.

The core idea is to use a Siamese Network~\cite{48e4505b7f124ee89368447f32b5cf0e}, a neural architecture that learns task-specific similarity by comparing (the representation of) input pairs, to compute a similarity score between generated and original graphs in the embedding space.
This allows us to assess how closely their learned representations align.
In fact, the generation process should be able to produce graphs that resemble the original ones in some way, and a non-trivial similarity function can help quantify this resemblance that can hardly be captured by MMD or other metrics, as discussed above.
The model could be trained, as common in deep learning, in an end-to-end fashion, using a set of labeled graphs, where the labels indicate whether the two graphs are similar or not.
\begin{figure*}[ht]
    \centering
    \includegraphics[width=0.99\textwidth]{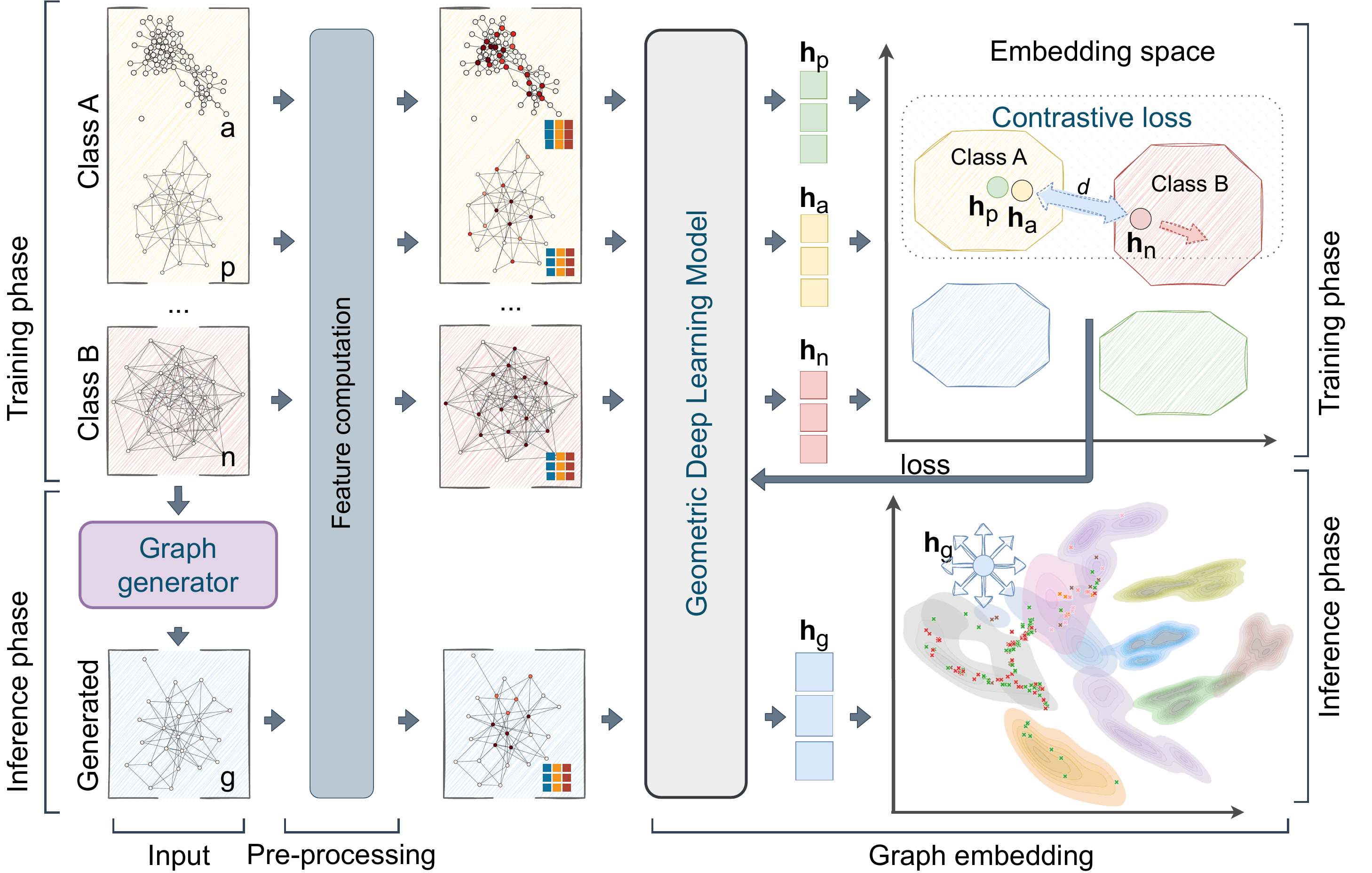}
    \caption{Graph similarity using our proposed method, RGM.
    RGM leverages a Siamese architecture, where a Graph Neural Network (GNN) --- specifically a Graph Attention Network (GAT) --- is trained to compute similarity between graphs.
    Input graphs are first pre-processed to extract informative node features (e.g., degree, $k$-core, local clustering coefficient, and chi-squared statistics), then passed through the GNN to obtain embeddings.
    During training, the model receives triplets of graphs: an anchor, a positive (same class), and a negative (different class).
    Their embeddings (e.g., $\mathbf{h}_a$, $\mathbf{h}_p$, and $\mathbf{h}_n$, respectively in the figure) are optimized using a triplet loss function that minimizes the distance between anchor and positive graphs, while maximizing the distance to the negative one.
    This process encourages the model to learn a similarity function that maps graphs into an embedding space where structurally similar graphs are close together and dissimilar ones are farther apart.
    After training, the model can evaluate generated graphs (e.g., $g$ in the figure) by comparing them to anchor graphs from the training set.
    It computes similarity scores and assigns the generated graph to the class most represented among its closest anchors.
    }
    \label{f:graph_similarity}
\end{figure*}

\section{Experiments}
This section presents the experimental setup and results of the classification task within our RGM framework, along with the training, generation, and evaluation procedures for the two selected GGMs: GRAN and EDGE.
The motivation behind selecting these two models lies in their distinct capabilities and generative approaches, which complement each other in capturing different aspects of graph structures. GRAN is an auto-regressive model that has demonstrated superior performance compared to GraphRNN~\cite{you2018graphrnngeneratingrealisticgraphs} in generating realistic graphs, particularly for datasets with complex structural dependencies. On the other hand, EDGE is a diffusion-based model that leverages iterative refinement to generate graphs, making it well-suited for capturing fine-grained details and global graph properties.
\subsection{Experimental Setup}
\paragraph{Dataset}
The experiments were conducted on a custom dataset drawn from the publicly available graph collection at the~\href{https://networkrepository.com}{\textit{Network Data Repository}}.
The dataset consists of graphs from five distinct classes: Biological, Connectome, Infrastructure, Internet, and Social.
To further assess the robustness of the model and extend its applicability across a broader range of graph types, synthetic graphs were incorporated into the dataset.
These synthetic graphs were generated using five well-known graph generation models: the Barabási-Albert model, the Erdős-Rényi model, the Stochastic Block Model, the Lancichinetti-Fortunato-Radicchi model, and the nonuniform Popularity-Similarity-Optimization model.
A total of $300$ graphs were generated for each model, featuring diverse statistics, using NetworkX~\cite{hagberg2008exploring} and iGraph~\cite{Csardi2006-qq} libraries.
This expanded dataset, encompassing $10$ classes in total, provides a comprehensive foundation for evaluating the performance of graph generative models across a wide range of applications and domains.

Due to the imbalanced class distribution in the dataset, we utilized the \href{https://github.com/ufoym/imbalanced-dataset-sampler}{ImbalancedDatasetSampler} from \textit{torchsampler} to create balanced data loaders.

For the first classification task, the dataset contains graphs with up to $|N| = 35{,}000$ nodes, with detailed statistics provided in the accompanying Table~\ref{t:dataset_info}.
However, for the experiments involving GGMs, the dataset size was reduced to graphs with at most $|N| = 3{,}000$ nodes to accommodate computational constraints.

\begin{table*}[ht]
    \centering
    \caption{Dataset statistics. %
    }
    \label{t:dataset_info}
    \rowcolors{3}{gray!20}{white}
    \begin{tabular}{llrrrrr}
        \toprule
        & \multirow{2}{*}{\textbf{Class}}
        & \multirow{2}{*}{\textbf{\#graphs}}
        & \multicolumn{2}{c}{\textbf{\#nodes}}
        & \multicolumn{2}{c}{\textbf{\#edges}} \\
        \cmidrule(lr){4-5}\cmidrule(lr){6-7}
        & & & Min & Max & Min & Max \\
        \midrule
        \textbf{Real} 
        & Biological   & 176 & 18  & 16\,656 & 60 & 1\,525\,644 \\
        & Connectome   & 530 & 19  & 21\,663 & 74 & 5\,558\,796 \\
        & Infrastructure & 313 & 16 & 21\,775 & 34 & 75\,894 \\
        & Internet     & 284 & 282 & 34\,904 & 2\,060 & 220\,922 \\
        & Social       & 264 & 17  & 32\,362 & 72 & 11\,148\,466 \\
        \midrule
        \textbf{Synthetic} 
        & BA    & 300 & 1\,000 & 1\,000 &  9\,950 & 15\,872 \\
        & ER    & 300 &   996  & 1\,000 &  8\,000 & 10\,000 \\
        & LFR   & 300 & 1\,000 & 1\,000 &  6\,190 & 18\,596 \\
        & nPSO  & 300 & 1\,000 & 1\,000 & 14\,026 & 17\,462 \\
        & SBM   & 300 & 1\,000 & 1\,000 & 14\,026 & 17\,462 \\
        \bottomrule
    \end{tabular}
\end{table*}

\subsection{Representation-aware Graph-generation Model
evaluation (RGM)}
The model was designed to produce graph embeddings that can be used for graph classification tasks.
In particular, it includes $3$ Graph Attention Network (GAT) layers --- a type of Graph Neural Network (GNN) that uses attention mechanisms to capture the importance of neighboring nodes in a graph and learn node embeddings --- with a hidden dimension of $8$ and $4$ attention heads.
The GNN layers are followed by a fully connected layer with a hidden dimension of $8$ and an output dimension equal to the number of classes in the dataset.

The input node features are computed using a set of topological features that are extracted from the graphs in the dataset.
Specifically, we use four input topological features that capture multiscale information about the graph structure: the node degree, a chi-square statistic computed over the node's neighborhood, the local clustering coefficient, and the k-core number.
While the node degree and chi-square statistic capture local connectivity patterns, the clustering coefficient and k-core number provide information about tightly connected groups of nodes.
For each graph, the node features are concatenated into a single feature vector, which is then fed into the GNN layers that propagate and aggregate them iteratively during the message-passing process.
For instance, when propagated, the node degree can be used to infer assortativity and heterogeneity in the graph structure, while the chi-square statistic can help identify local patterns and outliers.
The local clustering coefficient captures the tendency of nodes to cluster together locally, indicating the presence of tightly-knit communities or groups within the graph, while the k-core number provides information about the core structure of the graph.
The features, computed using graph-tool~\cite{peixoto_graph-tool_2014}, are leveraged to enrich the node representations and improve the model's ability to capture the underlying structure of the graphs, as shown in~\cite{grassia2021machine}.

The model was trained using a supervised approach, where the input consists of triplets of graphs: an anchor graph, a positive graph (same class as the anchor), and a negative graph (different class).
The framework is represented in Figure~\ref{f:graph_similarity}.
The training process was performed over $64\%$ of the dataset using a Triplet Margin Loss function, which encourages the model to learn embeddings that are closer for similar graphs and farther apart for dissimilar graphs, using the AdamW optimizer with a learning rate of $0.003$ and a weight decay of $5 \times 10^{-3}$;
 early stopping was used based on the validation loss.
The validation set consisted of $16\%$ of the total dataset, while the remaining $20\%$ of the graphs were reserved for testing.
Test graphs are classified into their respective classes using a balanced anchor selection strategy based on a dynamic $k$-NN approach.
Specifically, each generated graph is assigned to a class via a $k$-nearest neighbor classifier, which compares its embedding to those of the training anchors and selects the class of the closest ones.
The number of neighbors $k_i$ used for each class $i$ is determined dynamically as $k_i = \sqrt{|\mathrm{anchors}_i|}$, where $|\mathrm{anchors}_i|$ is the number of training anchors for that class, following the recommendation in~\cite{DBLP:journals/corr/HassanatAAA14}.

The final classification results are presented in the confusion matrix in Table~\ref{tab:final_cmall}, which details the classification performance on both real-world and synthetic graph classes.
The model achieves near-perfect classification performance on synthetic graph classes, consistently reaching $100\%$ accuracy for BA, ER, LFR, and nPSO, and $98\%$ for SBM.
This indicates that the model effectively captures the distinct structural signatures of graphs generated by canonical models, which often exhibit clear topological characteristics (e.g., clustering, degree distribution, modularity).
In contrast, performance on real-world graph classes remains strong but shows greater variability. The model attains perfect accuracy on the Internet class (100\%), and high accuracy on Connectome (92\%), Infrastructure (89\%), and Biological (83\%) graphs. However, classification of the Social class proves more challenging, with notable confusion: $19\%$ of Social graphs are misclassified as Infrastructure, $11\%$ as Biological, $9\%$ as Connectome, and $4\%$ as Internet.
This performance drop may reflect the inherently heterogeneous nature of social graphs, which often exhibit a mix of structural patterns, ranging from small-world to scale-free properties, which makes them more difficult to distinguish from other real-world networks.

\begin{table*}[ht]
\centering
\caption{Confusion matrix computed on the test set, including both real-world and synthetic graph classes.
}
\label{tab:final_cmall}
\rowcolors{2}{gray!25}{white}
\adjustbox{max width=\textwidth}{
\begin{tabular}{l|ccccc|ccccc}
\toprule
& \multicolumn{5}{c|}{\textbf{Real Classes}} & \multicolumn{5}{c}{\textbf{Synthetic Classes}} \\
\textbf{True$\downarrow$ / Pred$\rightarrow$} & Biological & Connectome & Infrastructure & Internet & Social & BA & ER & LFR & nPSO & SBM \\
\midrule
Biological     & \textbf{83.0} & 0.0 & 0.0 & 3.0 & 10.0 & 0.0 & 0.0 & 0.0 & 3.0 & 0.0 \\
Connectome     & 0.0 & \textbf{92.0} & 6.0 & 0.0 & 2.0 & 0.0 & 0.0 & 0.0 & 0.0 & 0.0  \\
Infrastructure & 0.0 & 3.0 & \textbf{89.0} & 3.0 & 5.0 & 0.0 & 0.0 & 0.0 & 0.0 & 0.0 \\
Internet       & 0.0 & 0.0 & 0.0 & \textbf{100.0} & 0.0 & 0.0 & 0.0 & 0.0 & 0.0 & 0.0 \\
Social         & 11.0 & 9.0 & 19.0 & 4.0 & \textbf{57.0} & 0.0 & 0.0 & 0.0 & 0.0 & 0.0 \\

\midrule
BA             & 0.0 & 0.0 & 0.0 & 0.0 & 0.0 & \textbf{100.0} & 0.0 & 0.0 & 0.0 & 0.0 \\
ER             & 0.0 & 0.0 & 0.0 & 0.0 & 0.0 & 0.0 & \textbf{100.0} & 0.0 & 0.0 & 0.0\\
LFR            & 0.0 & 0.0 & 0.0 & 0.0 & 0.0 & 0.0 & 0.0 & \textbf{100.0} & 0.0 & 0.0 \\
nPSO           & 0.0 & 0.0 & 0.0 & 0.0 & 0.0 & 0.0 & 0.0 & 0.0 & \textbf{100.0} & 0.0 \\
SBM            & 0.0 & 0.0 & 0.0 & 0.0 & 0.0 & 2.0 & 0.0 & 0.0 & 0.0 & \textbf{98.0} \\
\bottomrule
\end{tabular}
}
\end{table*}

\subsection{Graph Generation: Assessment with RGM}
\paragraph{GRAN Assessment}
For the experimental GRAN procedure, we followed the same setup and code as the original paper.
We conducted experiments using our dataset (without node features as they're not supported by the model) and the same training setup as in the original paper.
We note that, to aid the performance of GRAN (and EDGE), we performed separate training and generation for each class without mixing them.
The dataset was first divided per class, and then each class was split into training and test sets with a ratio of $80\%$ and $20\%$, respectively.
Sub-graph sampling was enabled to facilitate learning from large structures, and the model was trained on the training set for a maximum of $50,000$ epochs with periodic validation every $1000$ epochs via the MMD metrics, as per the original paper.
Following the GRAN default model, we employed the \textit{GRANMixtureBernoulli} model variant, configured with $20$ mixture components, $7$ GNN layers, and symmetric adjacency reconstruction.
Each graph was encoded using a $512$-dimensional hidden representation, with attention mechanisms active during message passing.
The model was trained using the Adam optimizer (learning rate: $0.0001$, $\beta_1 = 0.9$, $\beta_2 = 0.999$, no weight decay) with a batch size of $4$.
To maintain temporal consistency and preserve structure, a depth-first node ordering was used during generation.
The number of epochs for each class was determined through extensive trials, balancing computational feasibility and model performance.
The maximum graph size in terms of nodes was approximately $3{,}000$, which constrained the dataset size for both GRAN and EDGE generative models.

Upon completion of the training process, the next steps involved graph generation using the trained model and evaluation of the generated graphs.

The generated graphs were then classified into their respective classes using the pre-trained RGM framework combined with the $k$-nearest neighbors classifier with dynamic $k$.
We stress that we use the same networks to train the RGM and GRAN (and EDGE) models for fairness.

\begin{table*}[ht]
\centering
\caption{Classification confusion matrix for the graphs generated by the GRAN model.
}
\label{tab:final_cm_generated}
\rowcolors{2}{gray!25}{white}
\adjustbox{max width=\textwidth}{
\begin{tabular}{l|ccccc|ccccc}
\toprule
& \multicolumn{5}{c|}{\textbf{Real Classes}} & \multicolumn{5}{c}{\textbf{Synthetic Classes}} \\
\textbf{True$\downarrow$ / Pred$\rightarrow$} & Biological & Connectome & Infrastructure & Internet & Social  & BA & ER & LFR & nPSO & SBM \\
\midrule
Biological     & \textbf{83.0} & 0.0 & 13.0 & 0.0 & 3.0 & 0.0 & 0.0 & 0.0 & 0.0 & 0.0  \\
Connectome     & \textbf{97.0} & 0.0 & 0.0 & 1.0 & 1.0 & 0.0 & 0.0 & 0.0 & 0.0 & 0.0  \\
Infrastructure & \textbf{83.0} & 0.0 & 3.0 & 7.0 & 3.0 & 0.0 & 0.0& 0.0 & 0.0 & 3.0 \\
Internet       & \textbf{81.0} & 0.0 & 3.0 & 0.0 & 7.0 & 0.0 & 0.0 & 3.0 & 0.0 & 7.0  \\
Social         & 5.0 & 0.0 & 10.0 & 15.0 & \textbf{70.0} & 0.0 & 0.0 & 0.0 & 0.0 & 0.0   \\
\bottomrule
\end{tabular}
}
\end{table*}
The resulting confusion matrix, shown in Table~\ref{tab:final_cm_generated}, provides a quantitative measure of the generative model's performance in preserving class-specific graph characteristics.
In particular, we note that most of the graphs are misclassified, and often classified as biological.
For example, $83\%$ of the generated Biological graphs are classified correctly, while the remaining are classified as Social or Infrastructure.
On the other hand, $97\%$ of the generated Connectome graphs are classified as Biological.
Surprisingly, $70\%$ of the generated Social graphs are correctly classified as Social.
It appears that GRAN generates graphs with similar structures among them, regardless of the training graphs used for its training.
The behavior of the generated graphs appears problematic, indicating that GRAN struggles to produce graphs that effectively replicate the characteristics of the training data.
The only result that can be considered satisfactory is for the Social class.

\paragraph{EDGE Assessment}
In the EDGE experimental process, we used the same configuration as the original experiments, but, as the paper's authors suggest, we adjusted the number of \textit{diffusion steps} based on the number of edges in the dataset. 
The training process was designed to accommodate large graphs, with a maximum graph size of $3{,}000$ nodes, and the batch size was set to $2$ to manage memory constraints.
Importance sampling was employed to enhance the efficiency of graph generation, while the noise schedule was configured linearly to ensure a smooth diffusion process. These choices were made to ensure the model's ability to generate high-quality synthetic graphs that accurately reflect the structural properties of the original dataset. Key parameters included a diffusion dimensionality of $64$, Adam optimizer with a learning rate of $0.0001$, and a dropout rate of $0.1$. 

The EDGE model was trained for a variable number of epochs due to the high computational cost of the procedure, again on the same data (class by class) used to train the RGM framework.
To ensure efficient use of resources, we monitored the evaluation metrics and stopped training when performance plateaued or improvements became negligible.
For the evaluation, we adopted the same methodology used for GRAN: generated graph embeddings were compared against the training anchors using our RGM framework.
This consistent approach allows for a direct and meaningful comparison between the two generative models, highlighting their respective strengths and limitations in replicating the structural properties of the original graphs.

\begin{table*}[ht]
\centering
\caption{Classification confusion matrix for the graphs generated by the EDGE model.
}
\label{tab:cm_edge}
\rowcolors{2}{gray!25}{white}
\adjustbox{max width=\textwidth}{
\begin{tabular}{l|ccccc|ccccc}
\toprule
& \multicolumn{5}{c|}{\textbf{Real Classes}} & \multicolumn{5}{c}{\textbf{Synthetic Classes}} \\
\textbf{True$\downarrow$ / Pred$\rightarrow$} & Biological & Connectome & Infrastructure & Internet & Social & BA & ER & LFR & nPSO & SBM \\
\midrule
Biological     & \textbf{80.0} & 0.0 & 7.0 & 3.0 & 10.0 & 0.0 & 0.0 & 0.0 & 0.0 & 0.0   \\
Connectome     & \textbf{100.0} & 0.0 & 0.0 & 0.0 & 0.0 & 0.0 & 0.0 & 0.0 & 0.0 & 0.0 \\
Infrastructure & \textbf{70.0} & 0.0 & 3.0 & 0.0 & 17.0 & 10.0 & 0.0 & 0.0 & 0.0 & 0.0  \\
Internet       & \textbf{52.0} & 0.0 & 23.0 & 13.0 & 8.0 & 1.0 & 0.0 & 0.0 & 0.0 & 3.0  \\
Social         & 33.0 & 0.0 & 0.0 & \textbf{47.0} & 20.0 & 10.0 & 0.0 & 0.0 & 0.0 & 0.0   \\

\bottomrule
\end{tabular}
}
\end{table*}

\paragraph{MMD Assessment}
Here we present the results of the Maximum Mean Discrepancy (MMD) evaluation for both the GRAN and EDGE models, which should be interpreted as a measure of the quality of the generated graphs.
In fact, while we have introduced our own evaluation framework, MMD remains the standard metric widely used in the community to assess the quality of graph generative processes.
Therefore, for consistency and to align with existing literature, we include MMD computations as part of our evaluation.
This allows us not only to benchmark the performance of the generative models and position our findings within the broader scope of graph generation research, but also to stress that low MMD values of specific metrics do not necessarily correlate with the model's ability to generate graphs that are structurally similar to the original ones.

To compute the MMD, we utilized the implementation provided in the EDGE repository, which itself is derived from the original GraphRNN repository, widely adopted by the community in the context of graph generative models.
In line with standard practices, these metrics were computed by comparing the reference dataset, which was used to train the generative models, with the generated dataset.
Since the MMD captures the distance between two distributions, the lower the MMD value, the closer the two distributions are.

Here, we computed the MMD values for the following metrics: degree, clustering coefficient, Orbits, spectral, and Neighborhood Subgraph Pairwise Distance Kernel (NSPDK).
In particular, the degree and the clustering coefficient distributions are local properties of the nodes that capture the number of connections and the tendency of the node's neighbors to connect with each other, respectively.
The Orbits metric, computed using the Orca algorithm~\cite{hovcevar2014combinatorial}, captures the number of node- and link-orbits (i.e., the number of distinct ways a node can be connected to its neighbors) of 4- and 5-node graphlets in the network, providing a measure of the local structure around each node.
The spectral metric captures the eigenvalues of the adjacency matrix, which can provide insights into the global structure of the graph.
Finally, the NSPDK (Neighborhood Subgraph Pairwise Distance Kernel)~\cite{costa2010fast} metric captures the distance between pairs of nodes in the graph based on their neighborhood structure, providing a measure of the similarity between different parts of the graph.

\begin{table*}[ht]
\centering
\caption{MMD metric values for graphs generated by GRAN. Values are computed between the reference dataset and the generated graphs; lower values indicate better similarity.}
\centering
\rowcolors{2}{gray!25}{white}
\adjustbox{max width=\textwidth}{

\begin{tabular}{llllll}

\toprule
 \textbf{Class$\downarrow$ / MMD$\rightarrow$}    & Degree & Clustering & Orbits & Spectral & NSPDK  \\
 \midrule
 Biological &  0.087858  & 0.082313 & 0.045951 & 0.047899 & 0.044767 \\
 Connectome  & 0.039575 & 0.038097 & 0.015204 & 0.747242 & 0.014247     \\
 Infrastructure & 0.110093 & 0.010552 & 0.019894 & 0.029463 & 0.022458 \\
 Internet  & 0.369897 & 0.779779 & 0.035310 & 0.244249 & 0.040802\\
 Social & 0.043118 & 0.066998 & 0.042369 & 0.046709 & 0.050510\\
 \bottomrule
\end{tabular}
}
\label{t:gran_mmd}

\end{table*}

\begin{table*}[ht]
    \caption{MMD metric values for graphs generated by EDGE. Values are computed between the reference dataset and the generated graphs; lower values indicate better similarity.}
    \label{t:edge_mmd}
    \centering
    \rowcolors{2}{gray!25}{white}
    \adjustbox{max width=\textwidth}{
        \begin{tabular}{llllll}
            \toprule
            \textbf{Class$\downarrow$ / MMD$\rightarrow$}      &  Degree &
            Clustering & Orbits & Spectral & NSPDK  \\  
            \midrule
            Biological & 0.038625 & 0.405843 & 0.040677 & 0.046560 &  0.040510\\
            Connectome & 0.062112 & 0.873837 & 0.016176 & 1.146303 & 0.018985 \\
            Infrastructure & 0.065504 & 0.311149 & 0.021443 & 0.022558 & 0.031345 \\
            Internet & 0.291295 & 0.719936 & 0.035310 & 0.466615 & 0.036937 \\
            Social & 0.047823 & 0.104931 & 0.042346 & 0.050037 & 0.040986 \\
            \bottomrule
        \end{tabular}
    }
\end{table*}

In the MMD evaluation Tables~\ref{t:gran_mmd} and~\ref{t:edge_mmd}, the computed MMD values between the reference and generated datasets show mixed results even after extensive training, with most metrics indicating good alignment while others reveal significant discrepancies~\cite{you2018graphrnngeneratingrealisticgraphs,liao2020efficientgraphgenerationgraph}.
For instance, clustering coefficient MMD values are particularly high for Internet graphs (0.780 for GRAN, 0.720 for EDGE) and Connectome graphs (0.874 for EDGE), while spectral properties show substantial deviations, especially for Connectome graphs (1.146 for EDGE, 0.747 for GRAN).
In any case, despite most low MMD values (e.g., $< 0.05$ for several NSPDK and Orbits metrics), our RGM framework is unable to accurately classify the generated graphs.
Based on the previously reported results, we can assume that our model captures relevant structural differences.
Its poor performance in assigning the correct domain to generated graphs suggests that they still lack important characteristics of the original data, highlighting serious limitations of current state-of-the-art GGMs.
This discrepancy also highlights a key limitation of current usage of MMD: low values across certain statistics do not necessarily imply high structural fidelity or downstream utility.
Although individual metrics may appear well aligned, the overall similarity remains inadequate, as also shown by the statistical comparison of a wide set of network science metrics.
%
%
In fact, by looking at Figures A.1, A.2 and A.3
in the Appendix,
we can notice that there are substantial deviations in key topological properties when compared to the original graphs (i.e., the baseline used to train the GGMs).
For instance, both EDGE and GRAN struggle to accurately replicate assortativity, average clustering coefficient and transitivity, affecting the homophily and the formation of tightly-knit groups in the network.
The density, diameter and average path length are also misaligned, especially for GRAN.
The k-core analysis exhibits a significant mismatch in the Internet class for GRAN, and both GRAN and EDGE fail to capture the expected k-core structure in the Biological and Social classes.

\section{Conclusion}
%
In this work, we present a novel framework for evaluating the generative capabilities of Graph Generative Models (GGMs) using Geometric Deep Learning (GDL) techniques.
Our framework, RGM (Representation-aware Graph-generation Model evaluation), is based on a Siamese Graph Neural Network trained to recognize similar structural patterns in graphs, and is capable of classifying graphs into their respective domains based on their topological structure with high accuracy.

We applied this framework to evaluate the generative capabilities of two state-of-the-art GGMs, GRAN and EDGE, after training it on a custom dataset of real-world and synthetic graphs, to enhance the model's generalization and usage.
Our results demonstrate two critical findings: first, that current graph generative models struggle to produce graphs that effectively capture the structural characteristics that distinguish different graph domains; second, that Maximum Mean Discrepancy (MMD), the most commonly used evaluation metric in the field, fails to adequately assess the quality of generated graphs.

Specifically, while both GRAN and EDGE achieve low MMD values across various topological metrics, suggesting good distributional alignment, our RGM reveals significant limitations in their ability to preserve the essential structural properties that define different graph classes.
This discrepancy highlights the limitations of MMD as a sole evaluation metric and underscores the need for more sophisticated evaluation approaches.

Our findings have important implications for the graph generation community, suggesting that researchers should move beyond simple statistical measures and adopt more nuanced evaluation frameworks that can capture the complex structural properties that make graphs from different domains distinguishable.
For instance, the insights gained from this evaluation could inform the development of improved Graph Generative Models that explicitly optimize for preserving domain-specific structural characteristics rather than just statistical properties.




\section{Acknowledgment}
M.G. and G.M. acknowledge financial support from PNRR MUR project
CN00000013 – NATIONAL CENTRE FOR HPC, BIG DATA
AND QUANTUM COMPUTING. S.R. and G.M. acknowledges financial support from PNRR MUR project PE0000013-FAIR.
M.G. and G.M. acknowledge partial financial support from the University of Catania in the form of a PIAno di inCEntivi per la Ricerca di Ateneo (PIACERI). 

\bibliographystyle{IEEEtran}
\bibliography{bib}

\section{Biography Section}

\vspace{11pt}
\begin{IEEEbiography}
 [{\includegraphics[width=1in,height=1.25in,clip,keepaspectratio]{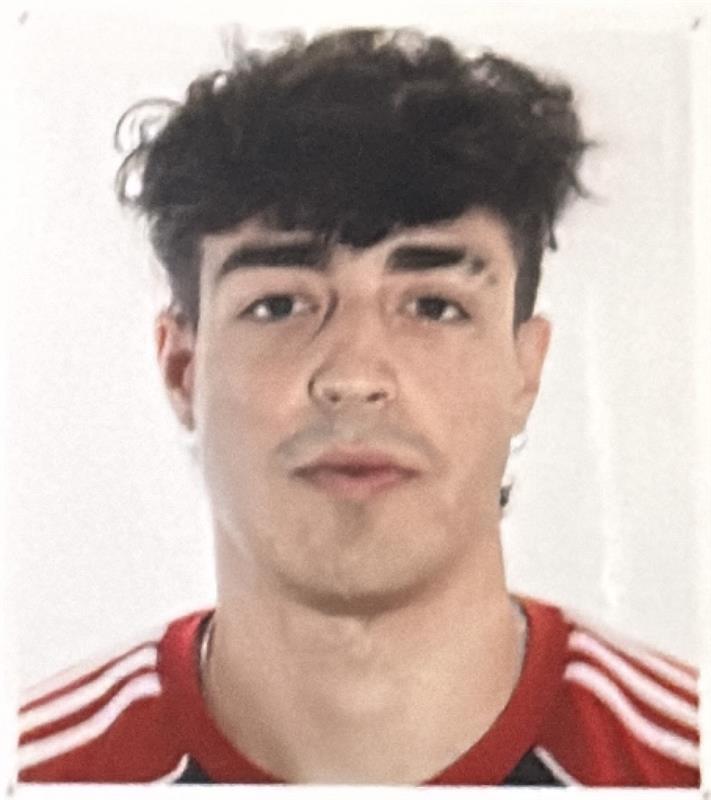}}]{Salvatore Romano} 
is a PhD student enrolled in the National PhD in Artificial Intelligence, XXXIX cycle, organized by University Campus Bio-Medico of Rome (Rome, Italy) and affiliated with University of Catania (Catania, Italy). He received his MSc in Data Science from the University of Catania. His research interests span from geometric deep learning to graph generation.
 \end{IEEEbiography}

 \begin{IEEEbiography}[{\includegraphics[width=1in,height=1.25in,clip,keepaspectratio]{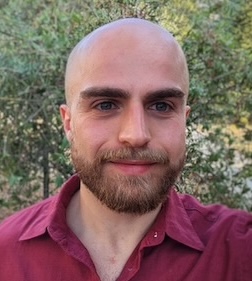}}]{Marco Grassia}
 is an Assistant Professor at the Department of Electrical, Electronic and Computer Engineering (DIEEI) of the University of Catania (Catania, Italy). He received his PhD in Computer Science from the University of Catania in 2018. His research interests include network science, geometric deep learning and their applications to complex systems.
 \end{IEEEbiography}

\begin{IEEEbiography}[{\includegraphics[width=1in,height=1.25in,clip,keepaspectratio]{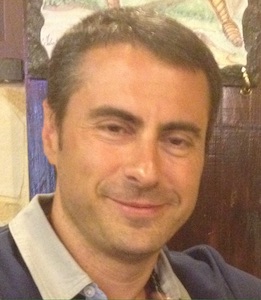}}]{Giuseppe Mangioni} 
is associate professor in Department of
 Electrical Electronics and Computer Engineering at the University of Catania.
 He received the degree in Computer Engineering (1995) and
 the Ph.D. degree (2000) at the University of Catania. Currently
 he teaches algorithms, internet and security and OOP at the University of Catania. His research interests include network science, machine learning and geometric deep learning. He is co-initiator and chair of the steering committee of the Int. Conference on Complex Networks (CompleNet).
\end{IEEEbiography}



\vfill

\end{document}


\title{Beyond MMD: Evaluating Graph Generative Models with Geometric Deep Learning - Appendix}

\author{%
Salvatore Romano~\orcidlink{0009-0002-2550-4435},
Marco Grassia~\orcidlink{0000-0001-5841-6058},
Giuseppe Mangioni~\orcidlink{0000-0001-6910-0112} \\
}


\markboth{Journal of \LaTeX\ Class Files,~Vol.~18, No.~9, September~2020}%
{Romano \MakeLowercase{\textit{et al.}}: Beyond MMD: Evaluating Graph Generative Models with Geometric Deep Learning}


\maketitle

 \begin{figure*}[ht]
    \centering
    \includegraphics[width=1.\textwidth,  page=1]{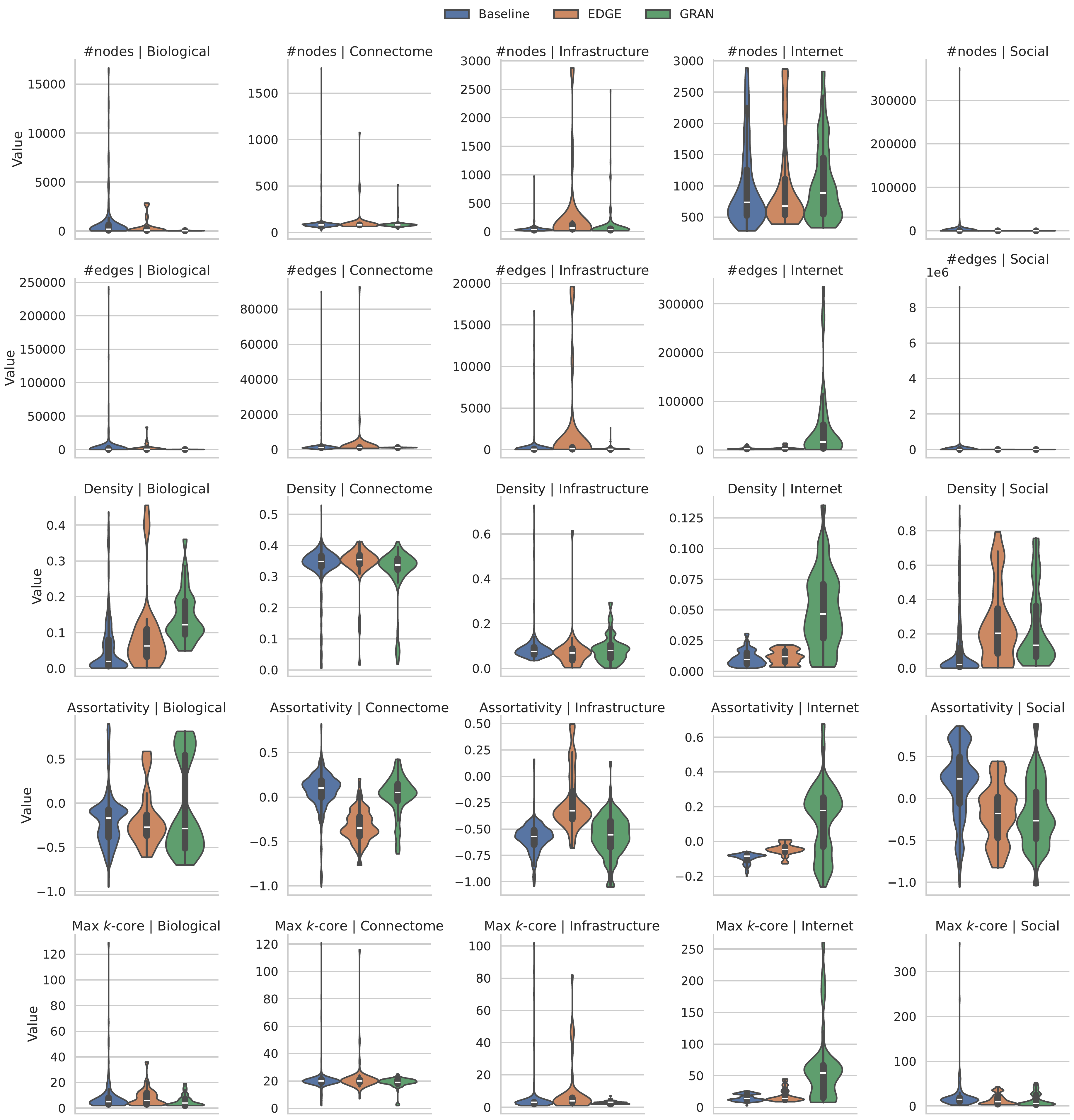}
    \caption{Comparison of topological properties between original and generated graphs. ``Assortativity'' is the degree assortativity coefficient, ``density'' is the density of the graph.}
    \label{f:topological-properties-1}
\end{figure*}

\begin{figure*}[ht]
    \centering
    \includegraphics[width=1.\textwidth, page=2]{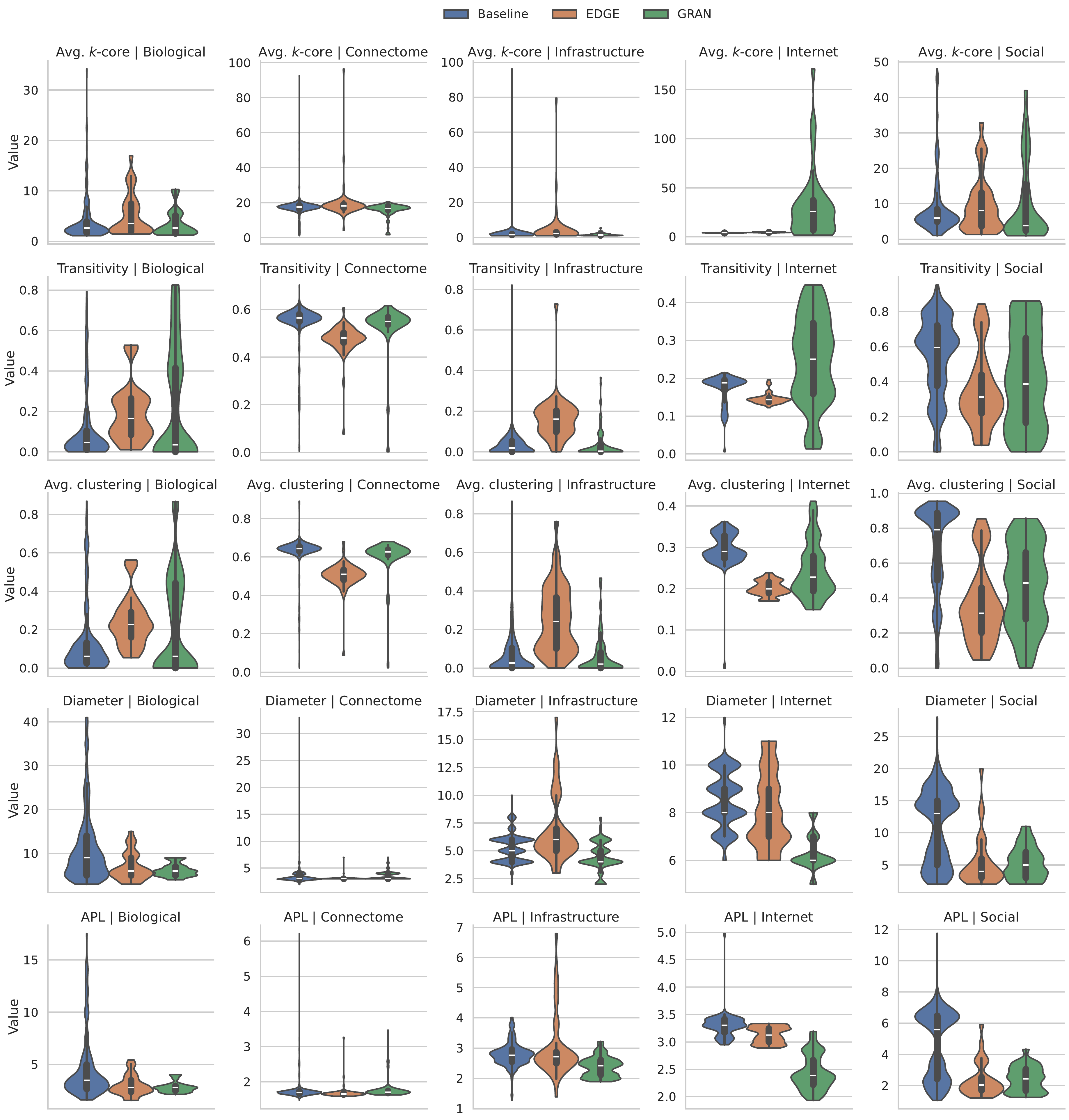}
    \caption{Comparison of topological properties between original and generated graphs. ``Clustering'' is the local clustering coefficient, ``diameter'' is the diameter of the graph,  ``APL'' is the Average Path Length.}
    \label{f:topological-properties-2}
\end{figure*}
\newpage
\begin{figure*}[ht]
    \centering
    \includegraphics[width=1.\textwidth, page=3]{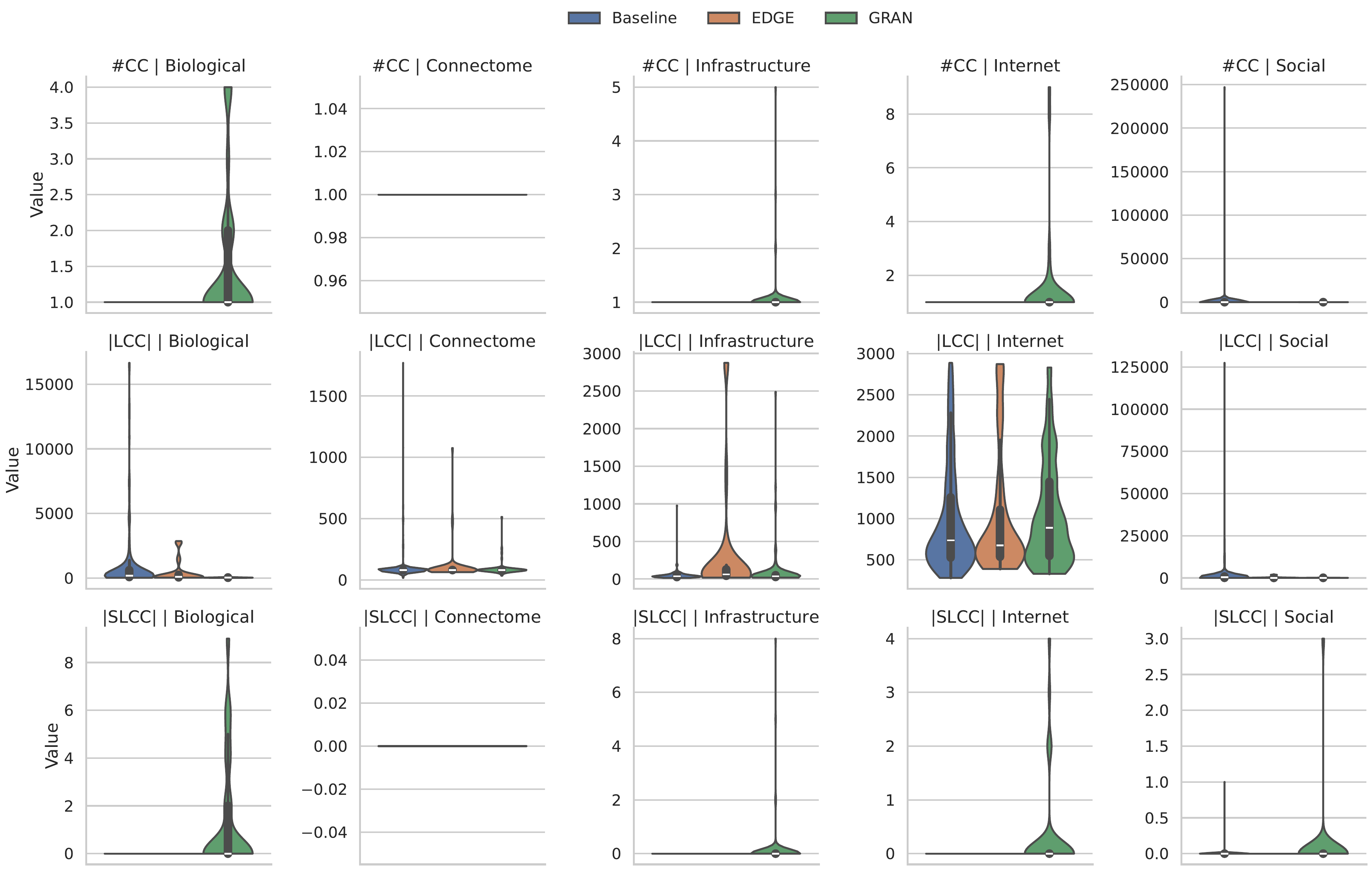}
    \caption{Comparison of topological properties between original and generated graphs. ``CC'' stands for Connected Components, ``$|\mathrm{LCC}|$'' is the size of the Largest Connected Component, ``$|\mathrm{SLCC}|$'' is the size of the Second Largest Connected Component.}
    \label{f:topological-properties-3}
\end{figure*}

\vfill